\documentclass{article}

\usepackage{arxiv}
\usepackage[numbers,sort&compress]{natbib}
\usepackage{authblk}

\usepackage{tabularx}
\usepackage{array}
\newcolumntype{Y}{>{\raggedright\arraybackslash}X}
\usepackage[utf8]{inputenc}
\usepackage[T1]{fontenc}
\usepackage{url}
\usepackage{booktabs}
\usepackage{amsfonts}
\usepackage{amsmath}
\usepackage{microtype}
\usepackage{xcolor}
\usepackage{enumitem}
\usepackage{tikz}
\usetikzlibrary{matrix,positioning}
\usepackage{wrapfig}
\usepackage[hypertexnames=false,breaklinks=true]{hyperref}
\usepackage{float}

\newcommand{\tabletight}{\footnotesize\renewcommand{\arraystretch}{1.1}\setlength{\tabcolsep}{4pt}}

\title{The Open-Box Fallacy: Why AI Deployment Needs\\ a Calibrated Verification Regime}

\author[1]{Phongsakon Mark Konrad\thanks{Corresponding author: \texttt{phkon23@student.sdu.dk}}}
\author[1]{Tim Lukas Adam}
\author[1]{Ane Cathrine Holst Merrild}
\author[1]{Riccardo Terrenzi}
\author[1]{Rebecca De Rosa}
\author[2]{Toygar Tanyel}
\author[1]{Serkan Ayvaz}
\affil[1]{Centre for Industrial Software, University of Southern Denmark, Alsion 2, S\o nderborg, 6400, Denmark}
\affil[2]{ProMake, Newark, DE, USA}

\begin{document}
\maketitle

\begin{abstract}
AI deployment in sensitive domains such as health care, credit, employment, and criminal justice is often treated as unsafe to authorize until model internals can be explained. This often leads to an excessive reliance on mechanistic interpretability to address a deployment challenge beyond its intended scope. We argue that the gate should instead be calibrated verification: authorization should be domain-scoped, independently checkable, monitored after release, accountable, contestable, and revocable. The reason is twofold. First, model capability is uneven across nearby tasks, so authorization must attach to a specific use rather than to a model in general. Second, societies have long governed opaque expertise through credentials, monitoring, liability, appeal, and revocation rather than mechanism-level explanation. Recent evidence reinforces this distinction between mechanistic understanding and deployment authority: a 53-percentage-point gap between internal representations and output correction shows that understanding may not translate into action, while one scoping review found that only 9.0\% of FDA-approved AI/ML device documents contained a prospective post-market surveillance study. We propose Verification Coverage, a six-component reportable standard with a minimum-composition rule, as the metric that should sit beside capability scores in model cards, leaderboards, and regulatory disclosures.
\end{abstract}

\section{Introduction}
\label{sec:intro}

AI systems are increasingly moving into operational settings where their outputs can affect access to care, economic opportunity, public services, physical security, or legal standing \citep{fda_ai_devices,cfpb_2022_03,state_v_loomis}. When these systems are deployed, trust is often sought through greater transparency and explainability \citep{doshiVelezKim2017,gilpin2018}, with the aim that understanding how a model reaches its output will make deployment decisions more justified, auditable, and safe \citep{nist_airmf,raji2020closing,mitchell2019model}. This is well motivated: mechanistic interpretability can expose safety-relevant internal structure and reveal failure modes \citep{anthropic2025biology,hubinger2024sleeper}. However, it does not by itself answer the deployment question.

Systems can be useful in domains where their internal mechanisms remain only partially understood \citep{fda_ai_devices,deFauw2018,dellacqua2023jagged}, and the question facing decision-makers is not whether opacity disappears before use, but what criteria are sufficient for deployment in a specific use case. This is reflected in the EU AI Act \citep{eu_ai_act_2024}, which regulates high-risk systems through deployment-facing obligations such as risk management, documentation, human oversight, and post-market monitoring. A credible answer must therefore distinguish model capability and functionality from deployment authority: evidence about how a system solves a task is not yet evidence that its use is appropriately scoped, monitored, and accountable.

We introduce the \emph{open-box principle} as the view that mechanistic transparency can provide important evidence for safer deployment, and the \emph{open-box fallacy} as the stronger demand that this evidence serve as the decisive deployment gate. Verification, as we use it, is a deployment-governance regime---formal checks, expert validation, institutional review, contestability, monitoring, and revocation \citep{nist_airmf,raji2020closing,wachter2017counterfactual}---that structures the evidence under which a deployment may be considered; final authority remains with the institutions named in the regime. The unit of authorization is a deployment in a domain, not a model in general: a freely distributed model produces many deployment contexts, each with its own deployer of record.

\textbf{This position paper argues that verifiers, not interpretability alone, should license AI deployment: mechanistic evidence is valuable, but deployment authority should rest on calibrated verification of a specific use.} Mechanistic evidence remains one stream within that regime, alongside behavioural evaluation, independent review, and stakeholder input.

The case for this position is twofold. First, model capability is uneven across nearby tasks, so authorization must attach to a specific use rather than to a model in general \citep{dellacqua2023jagged}. Second, opacity is not unique to AI deployment: mature institutions have long governed consequential expertise without complete mechanism-level access, relying instead on external checks and revision mechanisms. The question is therefore not whether mechanistic evidence matters, but whether the surrounding regime can make a particular deployment governable.

This paper makes three contributions:
\begin{itemize}[nosep, leftmargin=1.5em]
    \item We introduce the open-box fallacy and explain why it fails: model capability is uneven across nearby tasks, and institutional authorization has often operated under partial opacity.
    \item We develop calibrated verification as the alternative deployment regime, distinguishing evidence streams, verifier classes, and regime properties.
    \item We propose Verification Coverage as a six-component reportable standard with a minimum-composition rule, intended to sit beside capability scores in model cards, leaderboards, and regulatory disclosures.
\end{itemize}

\section{The Open-Box Principle and Fallacy}
\label{sec:fallacy}

The open-box principle holds that mechanistic evidence is genuinely deployment-relevant: understanding how a system produces its outputs informs whether and how it should be used. Internal inspection of large language models has been shown to recover causal structure---features and circuits that predict behavioural outcomes in ways surface statistics do not capture \citep{anthropic2025biology}. Mechanistic evidence can identify failure modes invisible to benchmark evaluation, distinguish genuine competence from spurious correlation, and inform the design of domain-specific tests.

The principle, correctly understood, supports a bounded inference: mechanistic evidence answers---partially and progressively---the question of how a system works, but not, by itself, whether its use is appropriately scoped, monitored, and accountable in a given context. The first question concerns explanation, the second authorization. The \emph{open-box fallacy} is the stronger inference that mechanistic evidence should be the decisive deployment condition. Mechanistic transparency is neither necessary when outputs can be independently checked, nor sufficient when a deployment lacks domain scope, monitoring, accountability, contestability, or revocation. A system may be internally opaque yet externally verifiable for a narrow deployment; conversely, a system may be mechanistically transparent yet inappropriate for deployment because it lacks those surrounding rails.

\subsection{Jagged capability across nearby tasks}
\label{sec:jagged}

A model's capability is not a single property that transfers smoothly from one use to another. Within the same workflow, the same system can improve performance on one task while degrading it on a seemingly similar task. This jagged capability profile is the central empirical fact that makes blanket deployment policy insufficiently granular. Categorical prohibition ignores domains where AI can outperform available alternatives; categorical authorization ignores domains where it remains unreliable. A jagged frontier motivates a granular response: authorization per domain, per verifier class, per use.

The ``jagged technological frontier'' captures this point: capability can reverse across a task boundary. In a field experiment with consultants, \citet{dellacqua2023jagged} find that AI improves productivity and quality on many consulting tasks, but reduces correctness on a complex managerial task beyond its demonstrated capability. On that task, consultants using AI are 19 percentage points less likely to produce correct solutions than those without AI. Model-level examples tell the same story: \citet{xu2025genius} call this the genius paradox, and \citet{fu2024counting} find that models recognize letters but fail to count them reliably, especially when letters repeat.

If AI were uniformly incompetent, deployment debates would be simple. If it were uniformly reliable, trust debates would also be simpler. The practical challenge is that models are capable enough to be useful, opaque enough to be hard to inspect, and uneven enough that neither blanket acceptance nor blanket refusal is adequate. The jagged frontier therefore motivates the paper's positive proposal. Authorization should attach to a deployment context, not to a model.

\subsection{Evaluation is already brittle}
\label{sec:eval-brittle}

The same problem appears in evaluation. If authorization should attach to a deployment context rather than to a model, then domain-blind benchmark scores are too coarse to carry the deployment gate by themselves. Frontier models often identify evaluation contexts: \citet{needham2025awareness} construct a benchmark of evaluation and deployment transcripts and report that Gemini~2.5~Pro reaches an AUC of $0.83$ at distinguishing them, against a human baseline of $0.92$. \citet{vanderweij2024sandbagging} show that language models can be prompted or trained to underperform selectively on dangerous-capability evaluations. 

Human-feedback evaluation is powerful but bounded: \citet{sharma2023sycophancy} show that preference judgments can reward sycophantic responses in some settings, and \citet{wen2024usophistry} show that, on complex QA and programming tasks, RLHF can make incorrect answers more convincing to human evaluators. The April 2025 rollback of a GPT-4o update after sycophantic behaviour reached production illustrates that this is not only a lab issue \citep{openai2025rollback}. Chain-of-thought rationales should not be treated as transparent windows either: \citet{chen2025reasoning} find that reasoning models often do not reveal when they used provided hints. Benchmarks and rationales are evidence, not authorization; the next section turns to the institutional conditions that make such evidence deployable.

\subsection{Institutional scaffolding under partial understanding}
\label{sec:trust}

Disciplined use under partial understanding has usually depended on instruments and institutions, not on complete mechanism-level explanation. The closest case is human expertise, but only when the expert is understood as part of an institutional surround rather than as an isolated mind. The analogy is institutional, not anthropomorphic. Societies authorize high-stakes human judgment without transparent access to the mental mechanism that generates each decision, but they do not authorize it blindly. They rely on training, examinations, supervised practice, professional registries, monitoring, liability, appeal, discipline, and revocation.

Authorizing a surgeon to practice does not require a theory of her cognition. It requires evidence of competence within a scoped role, monitoring of outcomes, malpractice exposure, and a procedure for withdrawing authority when warranted. Board exams and professional review may probe domain knowledge directly, including mechanisms of disease or treatment, but the object being authorized is not the neural process that generates each judgment. It is performance under institutional constraints. The institutional regime closes the loop that neither internal inspection nor outcome statistics closes alone. Mechanistic access may improve diagnosis, auditing, and oversight, but it is not the form in which authority is usually granted. Demanding mechanism-level understanding of AI as the universal authorization condition is therefore not a stricter version of standard practice; it is a different standard, applied to a different question.

Prior work notes versions of the asymmetry. \citet{zerilli2019transparency} argue that transparency demands placed on algorithmic systems can exceed what is demanded of human decision-makers; \citet{kempt2022relative} extend the comparison to clinical AI; \citet{jonas2017could} make vivid that even complete access to a system's parts does not guarantee functional understanding. The conclusion is not that AI should be treated as a human expert. It is that high-stakes authorization should be scoped and institutional rather than model-general and mechanism-dependent.

\section{Calibrated Verification}

This section turns the negative claim into a positive deployment regime. If authority attaches to a deployment context rather than to a model, the gate must ask whether that context is checkable, accountable, monitored, contestable, and revocable.

The concepts play different roles. Evidence streams describe where deployment-relevant information comes from. Verifier classes describe how outputs are checked. Regime properties describe what a deployment must provide before authority is granted. Verification Coverage, introduced in Section~\ref{sec:measure}, reports whether those properties are present.
\subsection{Four evidence streams}
\label{sec:streams}

\paragraph{Mechanistic evidence.} Internal inspection can reveal causal structure that behavioural evaluation misses \citep{anthropic2025biology}. It can sharpen accounts of failure modes, support anomaly detection, and provide distinctive evidence about internal objectives \citep{hubinger2024sleeper}. But evidence about internals is not the same as deployment authority. In one clinical setting, \citet{basu2026actionability} report a 53-point gap between detecting hazard information in the model’s internal representations and the model’s output behaviour. \citet{wu2025axbench} likewise find that, on steering benchmarks, prompting and finetuning outperform sparse-autoencoder methods. These results do not make mechanistic evidence irrelevant. Instead they show that it must be connected to population validity, post-release stability, accountability, and contestability before it can help authorize deployment.

\paragraph{Behavioural evaluation.} Held-out tests, red-team probes, prospective trials, and adversarial elicitation remain central. Scope limits include distribution shift, evaluation awareness \citep{needham2025awareness}, sandbagging \citep{vanderweij2024sandbagging}, sycophancy \citep{sharma2023sycophancy, wen2024usophistry}, and unfaithful rationales \citep{chen2025reasoning}. Behavioural evidence is necessary but not sufficient: a model may pass a benchmark and still drift after release, or pass a benchmark and still fail an affected person's contest.

\paragraph{Independent review.} Procedural audit, conflict-of-interest disclosure, and reproducibility checks examine the process that produced the evidence \citep{raji2020closing, mitchell2019model}. The reviewer may be a public regulator, accredited auditor, standards body, public-interest organization, or independent technical evaluator. Independent review does not substitute for empirical or mechanistic evidence. It checks whether those evidence streams were produced under conditions that resist self-serving framing.

\paragraph{Domain-expert and stakeholder input.} Affected parties detect harms, contestability gaps, and context-specific failure modes that evaluators miss. Counterfactual explanation \citep{wachter2017counterfactual} and adverse-action notice show that contestability is achievable under opacity. Stakeholder input should therefore be structured, documented, and tied to review criteria. It is not a substitute for evidence, nor a simple majority vote.

\subsection{Verifier classes} Evidence streams are not themselves verifiers. They become deployment-relevant when connected to checks. We distinguish three verifier classes. \emph{Formal verifiers} provide decisive checks inside a specified formal system: proof assistants, type checkers, cryptographic checks, and game rules. \emph{Empirical verifiers} provide probabilistic checks through observation: clinical trials, prospective validation, A/B tests, replication studies, crash statistics, field monitoring, and post-market surveillance. Test suites belong here, not with formal verifiers, since they are practical and high-coverage but incomplete. \emph{Social-normative verifiers} provide institutional checks for decisions that affect rights, opportunity, liberty, or legitimacy: credit, hiring, education, criminal justice, medicine, welfare, and public administration. The classes do not partition the four evidence streams. Each evidence stream can support different verifier classes. Section~\ref{sec:measure} maps these streams and classes onto the six Verification Coverage components.

\subsection{Plural verification and deployment accountability}
\label{sec:collective}

A verification regime is only as strong as its evaluators. Models that recognize evaluation contexts \citep{needham2025awareness}, sandbag selectively \citep{vanderweij2024sandbagging}, or sycophantically conform \citep{sharma2023sycophancy} can compromise empirical and human-mediated verifiers, just as they compromise benchmarks; formal verifiers remain decisive only within their specified formal system. The answer is not simply a more transparent model, but a more plural evaluation regime: a structured process in which independent technical evaluators, domain experts, affected stakeholders, and where safe public or open-source auditors evaluate a deployment context under predeclared criteria, with disagreement reported rather than averaged away and with named minority reports preserved when consensus is not reached.

Collective verification should not be understood as simple majoritarian judgment. Three properties make it operationally distinct: evaluators are scoped to the failure modes they are competent to detect; harnesses, rubrics, and failure taxonomies are open while sensitive tests, private data, and dual-use materials are withheld; final responsibility remains with named institutions rather than with the model or the metric. Verification Coverage supplies reportable necessary conditions for deployment; it does not replace institutional judgment about residual risk, public purpose, or domain-specific acceptability. Passing the threshold permits consideration rather than compels deployment, while failure on a required component blocks authorization unless the deployment is redesigned and re-evaluated.

\paragraph{Deployer of record.} Authority must attach to a named deployment context, with a primary accountable actor and role-specific duties distributed across the value chain. A freely distributed model therefore produces many separate deployment contexts, not one general authorization. Verification duties travel with the integrator into health care, credit, employment, criminal justice, or other consequential workflows; the same base model may be acceptable for low-stakes drafting and prohibited for autonomous diagnosis or criminal-risk scoring without contestability. Verifiers can still fail; what plural evaluation buys is an auditable trail when they do.

\subsection{From failure modes to six properties}
\label{sec:derive}

The six properties named in the position statement are not arbitrary. Each responds to a deployment failure mode that the preceding sections document, summarized in Table~\ref{tab:derive}.

\begin{table}[t]
\centering
\tabletight
\begin{tabular}{ll}
\toprule
Deployment problem & Required regime property \\
\midrule
Capability varies across nearby tasks (\S\ref{sec:jagged}) & Domain-scoped authorization \\
Self-reported evidence may be incomplete or conflicted (\S\ref{sec:streams}) & Independent checkability \\
Performance can change after release (\S\ref{sec:domains}) & Monitoring \\
Harm requires a responsible actor (\S\ref{sec:trust}) & Accountability \\
Affected parties detect errors evaluators miss (\S\ref{sec:streams}) & Contestability \\
New evidence should change deployment status (\S\ref{sec:collective}) & Revocation \\
\bottomrule
\end{tabular}
\caption{Calibrated verification derives its six properties from documented deployment failure modes. Each property answers a distinct failure mode and is non-compensable; strong evidence on one does not offset absence of another in a high-stakes domain.}
\label{tab:derive}
\end{table}

The properties are not interchangeable. A deployment with strong benchmark performance but no contest path cannot be treated as verified for a rights-affecting domain. A deployment with strong interpretability evidence but no monitoring plan is not adequately verified for a drifting clinical environment. A deployment with full auditing but no revocation trigger leaves no path to act when evidence accumulates. This is why Verification Coverage uses a minimum-composition rule rather than a weighted average.

\section{Verification Coverage}
\label{sec:measure}

A calibrated verification regime needs measurements. The field reports many capability scores, but it lacks a headline report of deployment verifiability. Verification Coverage is a reportable deployment-level metric in the minimal sense of a structured measurement object: a six-component profile with a minimum-composition rule. It is not a validated universal scalar score. The six components correspond one-to-one to the regime properties:

\begin{itemize}[nosep, leftmargin=1.5em]
    \item \textbf{Domain Coverage.} What fraction of real use falls inside the authorized domain? Proxy: log-share of queries inside declared scope.
    \item \textbf{Verifier Strength.} Are outputs formally, empirically, or institutionally checkable, and does evaluation quality hold as the model approaches the overseer's capability \citep{engels2025scaling}? Proxy: per-output presence of a check, plus capability-gap-conditioned evaluator-accuracy curves.
    \item \textbf{Monitoring Maturity.} Are post-deployment failures and drift detected? Proxy: presence of a statistically valid surveillance plan in the sense of \citet{dolin2025}.
    \item \textbf{Accountability Clarity.} Is a named actor responsible for the deployment? Proxy: identifiable accountable party in the deployment record.
    \item \textbf{Contestability.} Can affected people obtain reasons and appeal? Proxy: documented contest path with measured response time and outcome distribution.
    \item \textbf{Revocation Readiness.} Are there predeclared triggers and procedures for restricting or withdrawing deployment? Proxy: documented thresholds, incident triggers, named decision-maker, and time-to-action after a threshold breach.
\end{itemize}

For a deployment $d$, let $v(d)=(v_1(d),\ldots,v_6(d))\in\{0,1\}^6$ record whether each component is present and sufficiently documented for the deployment. Verification Coverage is reported first as this six-component profile, which makes explicit which rails of the regime are present and which are missing. The minimum-composition rule treats the weakest required rail as binding: $\mathrm{VC}_{\min}(d)=\min_i v_i(d)$, and a zero on any required rail withholds authorization for that deployment.

A scalar summary $\mathrm{VC}_{w,\mathcal{D}}(d)=\sum_i w_{i,\mathcal{D}}\,v_i(d)$, with non-negative weights summing to one and specified by the relevant evaluator or authorizing institution, may be reported for convenience but must never replace the profile: the same aggregate can hide different verification patterns. Table~\ref{tab:profile} illustrates this across existing regimes---medicine, credit, employment, autonomous systems, and criminal justice show different verification strengths and gaps across the six rails. Verification Coverage makes those differences reportable rather than collapsing them into model capability or functionality.

\section{Existing Domain Patterns and Gaps}
\label{sec:domains}

Existing high-stakes regimes already distinguish model output from deployment authority. Outputs that are unreviewed and automated, which affect health, safety, livelihood, rights, liberty, or irreversible action, are treated as not sufficient. The relevant instruments differ by domain. Some regimes emphasize authorization and review, others reason and appeal, and others monitoring and incident reporting. The common pattern is not full verification. It is a partial verification with uneven coverage across the six rails.

\paragraph{Scoring rule.}
Table~\ref{tab:profile} applies the six Verification Coverage components from Section~\ref{sec:measure}. The coding is diagnostic, not a legal compliance
determination. A cell is marked \textbf{H} when the component is instantiated by law, regulation, binding agency process, or routine public practice; \textbf{M} when it is partial, voluntary, indirect, or unevenly enforced; and \textbf{L} when it is thin, practically inaccessible, or not specifically attached to the AI deployment. The table scores the surrounding regime, not the intrinsic quality of
any particular model.

\paragraph{Medicine.} 
Medicine has strong domain-scoped authorization and review ability. For example, the FDA clinical decision support guidance asks whether software enables a healthcare professional to independently review the basis for recommendations so the professional does not rely primarily on the software \citep{fda_cds_guidance}. Additionally, FDA maintains a public list of AI-enabled medical devices authorized for marketing, while noting that the list is not comprehensive \citep{fda_ai_devices}. Clinical AI reporting guidelines such as CONSORT-AI, SPIRIT-AI, and DECIDE-AI likewise treat intended use, human interaction,
evaluation context, and error analysis as reportable objects \citep{liu2020consortai,rivera2020spiritai,vasey2022decideai}. The weak rail is post-release monitoring: a 2024 scoping review found that only 62 of 692 FDA-approved AI/ML devices (9.0\%) contained a prospective study for post-market surveillance \citep{muralidharan2024scoping}; \citet{dolin2025} use this gap to argue for statistically valid post-deployment monitoring. Medicine, therefore, shows both sides of the argument: authorization and reviewability can exist under partial mechanistic opacity, but lifecycle
monitoring remains incomplete.

\paragraph{Credit.}
Credit law has comparatively strong instruments of reason-giving and correction.
ECOA and Regulation B require specific reasons for adverse action; a 2022 CFPB circular stated that this requirement applies equally to credit decisions using complex and black-box credit
algorithms, although the circular was later withdrawn as guidance \citep{cfpb_2022_03,cfpb_withdrawal_2025}. The important point is the institutional pattern rather than the circular alone: consequential credit decisions must be tied to reasons, records, and a responsible creditor. Credit
therefore scores high on domain coverage, accountability, and contestability, but weaker on AI-specific monitoring and revocation. A denied applicant may receive reasons, but the public usually cannot see whether a deployed credit model is
drifting, how often appeals change outcomes, or what event withdraws the system from use.

\paragraph{Employment.}
Employment has a thinner audit-and-notice pattern. New York City's Local Law 144 prohibits the use of an automated employment decision tool unless it
has undergone a recent bias audit, a public summary is available, and required notices have been provided to candidates or employees \citep{nyc_ll144}. This is an independent review, but not necessarily individual contestability. A bias audit can report group-level disparity without giving a rejected applicant a meaningful
route to challenge the tool's role in the decision or to trigger suspension when a deployment fails. Employment therefore motivates the distinction between audit, notice, contestability, and revocation.

\paragraph{Autonomous systems and critical infrastructure.} Autonomous systems show a different profile: stronger incident reporting and
monitoring, weaker individual contestability. NHTSA's Standing General Order requires covered manufacturers and operators to report specified crashes involving Automated Driving Systems and Level 2 Advanced Driver Assistance Systems \citep{nhtsa_sgo_2025}. NHTSA's AV STEP proposal would add a voluntary review and reporting framework for certain ADS-equipped vehicles \citep{nhtsa_avstep}. The EU AI Act similarly classifies specified AI systems in
critical infrastructure, road traffic, employment, credit, education, and law enforcement as high-risk, and requires providers to establish post-market monitoring systems for high-risk AI
systems \citep{eu_ai_act_2024}. These instruments check deployed performance without requiring mechanism-level access. The thin rail is contestability for affected non-operators, such as pedestrians, passengers, and communities exposed
to system-level risk.

\paragraph{Criminal justice.}
Criminal justice illustrates low Verification Coverage in a high-stakes setting. In \emph{State v.\ Loomis}, the Wisconsin
Supreme Court allowed consideration of COMPAS with limitations and cautions \citep{state_v_loomis}. Subsequent work found that COMPAS did not outperform
a simple two-feature model or crowd-aggregated lay judgments on the Broward County recidivism-prediction task \citep{dressel_farid_2018}, and related scholarship has criticized the secrecy and procedural unfairness of COMPAS and similar proprietary recidivism risk tools \citep{rudin_wang_coker_2020}.
Criminal justice may have a named institutional decision-maker, but it lacks strong external verification, monitoring, and contestability when proprietary risk scores enter bail, sentencing, or parole workflows. For Verification Coverage, this is the limiting case: higher stakes require stronger verifiability, not simply stronger explanation demands.

\paragraph{Cross-domain pattern.}
The table shows why a single capability score cannot answer the deployment question. Medicine has authorization and reviewability but weak surveillance; credit has reasons and correction pathways but weak lifecycle monitoring; employment has audit and notice, but weak revocation; autonomous systems have incident reporting but weak appeal paths for affected non-operators; criminal
justice has named institutional authority, but weak verification and contestability. The minimum-composition rule turns this comparison into a
diagnosis: $\mathrm{VC}_{\min}(d)$ is set by the weakest rail, and the weakest
rail identifies the next repair.

\begin{table}[t]
\centering
\tabletight
\setlength{\tabcolsep}{4pt}
\begin{tabularx}{\textwidth}{Ycccccc}
\toprule
Regime & Domain & Verifier & Monitor. & Account. & Contest. & Revocation \\
\midrule
Medicine (FDA CDS; AI/ML devices)            & H & H & L & H & M & M \\
Credit (ECOA/Reg.~B; adverse action)         & H & M & L & H & H & M \\
Employment (NYC LL144; state AI laws)        & H & M & M & M & M & L \\
Autonomous systems / critical infrastructure & M & M & H & M & L & M \\
Criminal justice (post-\emph{Loomis})        & M & L & L & H & L & L \\
\bottomrule
\end{tabularx}
\caption{Illustrative profile of selected high-stakes regimes on the six
Verification Coverage components. Cells use the rule: \textbf{H} = component is
instantiated by statute, regulation, binding agency process, or routine domain
practice; \textbf{M} = partial, voluntary, indirect, or unevenly enforced;
\textbf{L} = thin, practically inaccessible, or not specifically attached to the
AI deployment. The table is a diagnostic application of the VC vocabulary, not an
empirical measurement or legal compliance determination.}
\label{tab:profile}
\end{table}

\section{Alternative Views}
\label{sec:alt}

We address seven objections, each either a partial concession, a misreading of the proposal, or an independent reason to accept it.

\paragraph{``General-purpose AI cannot be domain-gated.''}
General-purpose training does not entail general-purpose deployment authority. The jagged capability profile in Section~\ref{sec:jagged} makes blanket authorization too coarse in both directions: neither categorical prohibition nor categorical acceptance tracks a frontier that reverses across task boundaries \citep{dellacqua2023jagged}. Authority attaches to the deployer of record in a specific context; open distribution multiplies deployment contexts, not authorization.
\paragraph{``Some domains have no verifier.''}
That is an intended implication of the framework, not a failure of it. A domain without a plausible verifier class registers a low Verifier Strength score; the minimum-composition rule withholds authorization at that rail regardless of other scores. Decision support under a named accountable human remains available; autonomous consequential use does not.
\paragraph{``Open models make deployment gates unenforceable.''}
Open distribution shifts, rather than eliminates, the locus of duty. Verification obligations attach to whoever integrates a model into a consequential workflow. Enforcement is harder to achieve, but the same artifact produces many distinct deployment contexts, each with its own deployer of record and verification duties. The policy response is to clarify integration-layer accountability, not to treat availability as authorization.
\paragraph{``Interpretability is necessary for detecting deceptive internal objectives.''}
We accept the premise and contest the conclusion. Mechanistic inspection is plausibly the most direct signal for deceptive internal objectives \citep{hubinger2024sleeper}, and the regime weights it accordingly within the Mechanistic Evidence stream. The concession ends there: a 53-point gap between detecting hazard-relevant internal representations and correcting output behaviour \citep{basu2026actionability} shows that understanding does not imply control. Mechanistic evidence is necessary where it is the uniquely reliable signal; it is not sufficient as a universal gate.
\paragraph{``Internal auditing is enough'' \citep{raji2020closing}.}
Auditing is necessary but incomplete. It does not specify an external verifier class, an individual contestability path, or a predeclared revocation trigger. The employment row of Table~\ref{tab:profile} is instructive: NYC Local Law 144 mandates a bias audit yet leaves no meaningful appeal route for a rejected applicant and no revocation condition when disparity persists post-audit \citep{nyc_ll144}. Verification Coverage names what auditing leaves open.
\paragraph{``This repackages NIST or the EU AI Act.''}
Existing frameworks establish that AI assurance is multi-dimensional \citep{nist_airmf}; the EU AI Act and FDA guidance already require pieces of monitoring and documentation \citep{eu_ai_act_annex3, fda_cds_guidance}. The contribution is different: naming deployment verifiability as a missing \emph{reportable} object beside capability scores, and imposing a minimum-composition rule that makes the weakest rail binding. That gap is empirically visible: only nine percent of FDA-registered AI/ML devices include a prospective surveillance plan \citep{dolin2025}, a deficiency that would directly block authorization under the Monitoring Maturity floor.
\paragraph{``Verification Coverage will create false precision.''}
Reporting the six-component profile alongside any scalar prevents the aggregate from obscuring the weakest rail. The minimum-composition rule is itself anti-precision-inflating: it cannot be gamed by strengthening other components. The remaining safeguards are structural: independent assessment, preserved evaluator disagreement, and revocation when a verifier fails. False precision is an argument for these safeguards, not against measurement.

\section{Conclusion}
\label{sec:conclusion}

Societies have long authorized opaque expertise through training, credentials, monitoring, liability, appeal, and revocation, and mechanism-level access has often not been the primary basis for authorization. Model capability is uneven enough across nearby tasks that no blanket policy can cohere with that practice. Calibrated verification draws on four evidence streams---mechanistic, behavioural, independent, and stakeholder---and yields six regime properties: domain-scoped, independently checkable, monitored, accountable, contestable, and revocable. Verification Coverage reports those properties beside capability scores under a minimum-composition rule that names the weakest rail rather than averaging it out. The field should report not only what models can do, but also the verification regime that makes a specific deployment governable.

\bibliographystyle{plainnat}
\bibliography{references}

\end{document}